\documentclass[10pt,twocolumn,letterpaper]{article}

\usepackage{iccv}
\usepackage{times}
\usepackage{epsfig}
\usepackage{graphicx}
\usepackage{amsmath}
\usepackage{amssymb}
\usepackage{multirow}
\usepackage{adjustbox}
\usepackage{marvosym}

\usepackage{wrapfig}
\usepackage{diagbox}
\usepackage{epsfig}
\usepackage{extpfeil}
\usepackage{makecell}
\usepackage{colortbl}
\usepackage{float}
\usepackage{wrapfig}
\usepackage{booktabs}

\usepackage[pagebackref=false,breaklinks=true,colorlinks,bookmarks=false,citecolor=blue,linkcolor=blue]{hyperref}

\usepackage[capitalize]{cleveref}
\usepackage{algorithm}
\usepackage{algorithmic}

\usepackage{tabulary,overpic,xcolor}

\usepackage{color}

\usepackage{diagbox}

\definecolor{color3}{rgb}{0.95,0.95,0.95}
\definecolor{color4}{rgb}{0.96,0.96,0.86}
\definecolor{color5}{rgb}{0.90,0.90,0.90}

\iccvfinalcopy % *** Uncomment this line for the final submission

\def\iccvPaperID{2364} % *** Enter the ICCV Paper ID here

\ificcvfinal\fi

\begin{document}

%%%%%%%%% TITLE
\title{Retinexformer: One-stage Retinex-based \\ Transformer for  Low-light Image Enhancement}

\iffalse
\author{%
	Yuanhao Cai $^{1,2}$, Hao Bian $^{1,2,}$, Jing Lin $^{1,2}$, \\ Haoqian Wang $^{1,2,}$\textsuperscript{\Letter}, Radu Timofte $^{3,4}$, Yulun Zhang $^4$\\
	$^{1}$ Shenzhen International Graduate School, Tsinghua University, \\ $^2$  Shenzhen Institute of Future Media Technology, $^3$ University of W\"urzburg, $^4$ ETH Z\"{u}rich
}
\fi

\iffalse
\author{%
	Yuanhao Cai $^{1,2}$, Hao Bian $^{1,2}$, Jing Lin $^{1,2}$, \\ Haoqian Wang $^{1,2,~}$\thanks{Haoqian Wang is the corresponding author.}, Radu Timofte $^{3,4}$, Yulun Zhang $^4$\\
	$^{1}$ Shenzhen International Graduate School, Tsinghua University, \\ $^2$  Shenzhen Institute of Future Media Technology, $^3$ University of W\"urzburg, $^4$ ETH Z\"{u}rich
}
\fi

\author{%
	Yuanhao Cai $^{1}$, Hao Bian $^{1}$, Jing Lin $^{1}$, \\ Haoqian Wang $^{1,}$\thanks{Haoqian Wang and Yulun Zhang are corresponding authors}~~, Radu Timofte $^{2}$, Yulun Zhang $^{3, *}$\\
	$^{1}$ Tsinghua University, $^2$ University of W\"urzburg, $^3$ ETH Z\"{u}rich
}

\maketitle
\ificcvfinal\thispagestyle{empty}\fi

%%%%%%%%% ABSTRACT
\begin{abstract}
\vspace{-2mm}
When enhancing low-light images, many deep learning algorithms are based on the Retinex theory. However, the Retinex model does not consider the corruptions hidden in the dark or introduced by the light-up process. Besides, these methods usually require a tedious multi-stage training pipeline and rely on convolutional neural networks, showing limitations in capturing long-range dependencies. In this paper, we formulate a simple yet principled One-stage Retinex-based Framework (ORF). ORF first estimates the illumination information to light up the low-light image and then restores the corruption to produce the enhanced image. We design an Illumination-Guided Transformer (IGT) that utilizes illumination representations to direct the modeling of non-local interactions of regions with different lighting conditions. By plugging IGT into ORF, we obtain our algorithm, Retinexformer. Comprehensive quantitative and qualitative experiments demonstrate that our Retinexformer significantly outperforms state-of-the-art methods on thirteen benchmarks. The user study and application on low-light object detection also reveal the latent practical values of our method. Code is available at  \url{https://github.com/caiyuanhao1998/Retinexformer}
\end{abstract}

%%%%%%%%% BODY TEXT
\vspace{-3mm}
\section{Introduction}
\vspace{-1mm}
% intro of low-light enhancement
Low-light image enhancement is an important yet challenging task in computer vision. It aims to improve the poor visibility and low contrast of low-light images and restore the corruptions (\emph{e.g.}, noise, artifact, color distortion, \emph{etc.}) hidden in the dark or introduced by the light-up process. These issues challenge not only human visual perception but also other vision tasks like nighttime object detection.

\begin{figure}[htp]
	\begin{center}
		\begin{tabular}[t]{c} \hspace{-5.8mm} 
			\includegraphics[width=0.50\textwidth]{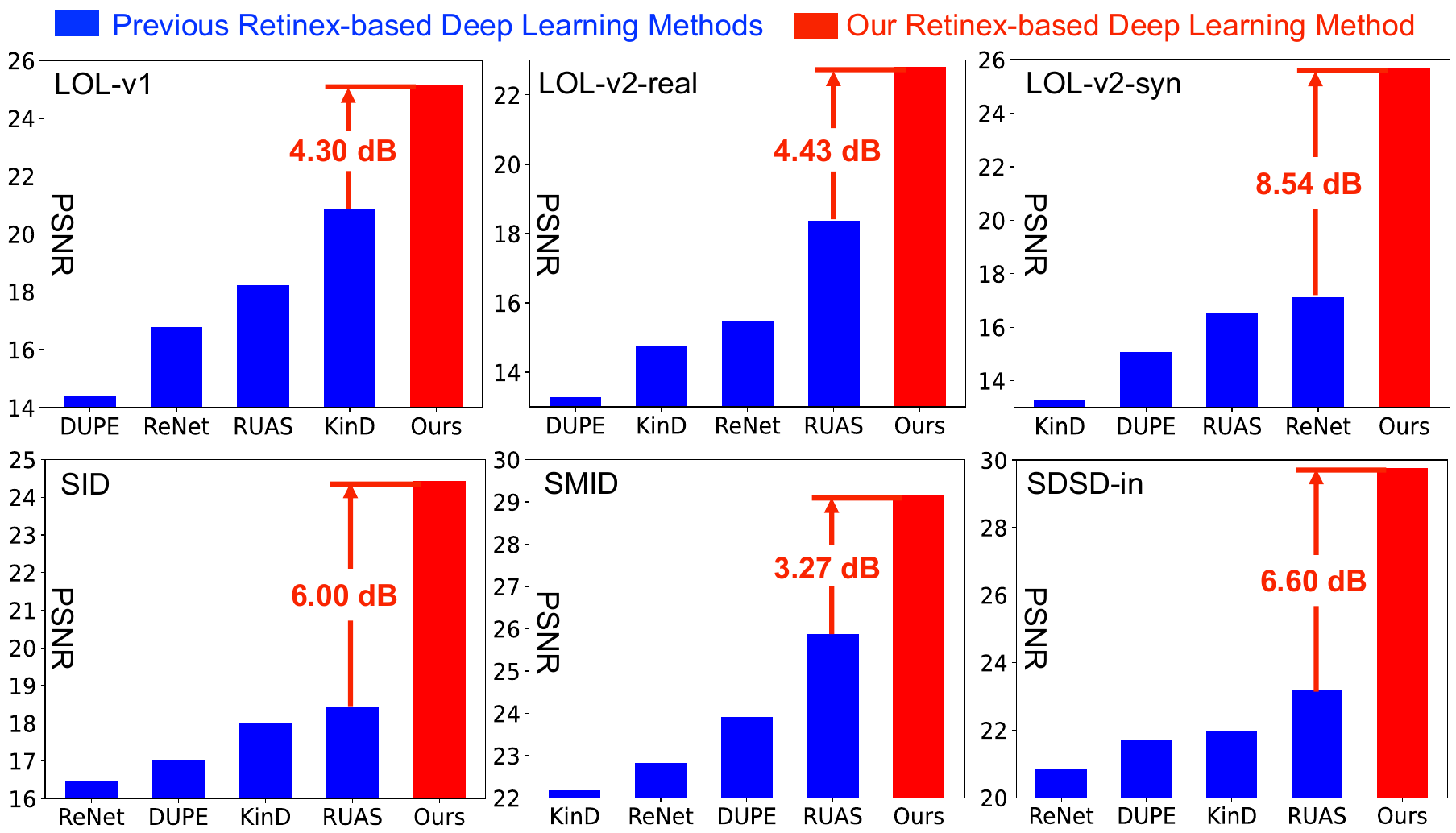}
		\end{tabular}
	\end{center}
	\vspace{-2mm}
	\caption{\small Our Retinexformer significantly outperforms state-of-the-art Retinex-based deep learning methods including DUPE (DeepUPE~\cite{deep_upe}), ReNet (RetinexNet~\cite{retinex_net}), KinD~\cite{kind}, and  RUAS~\cite{ruas} on six  low-light image enhancement benchmarks.}
	\label{fig:teaser}
	\vspace{-3mm}
\end{figure}

Hence, a large number of algorithms have been proposed for low-light image enhancement. However, these existing algorithms have their own drawbacks. Plain methods like histogram equalization and gamma correction tend to produce undesired artifacts because they barely consider the  illumination factors. Traditional cognition methods rely on the Retinex theory~\cite{retinex} that assumes the color image can be decomposed into two components, \emph{i.e.}, reflectance  and illumination. Different from plain methods, traditional methods  focus on illumination estimation but usually introduce severe noise or distort color locally because these methods assume that the images are noise- and color distortion-free. This is inconsistent with real under-exposed scenes. %Besides, these conventional Retinex-based algorithsms adopt hand-crafted image priors and need careful parameter tweeking, which usually requires a long time.

With the development of deep learning, convolutional neural networks (CNNs) have been applied in low-light image enhancement. These CNN-based methods are mainly divided into two categories. The first category directly employs a CNN to learn a brute-force mapping function from the low-light image to its normal-light counterpart, thereby ignoring human color perception. This kind of methods lack interpretability and theoretically proven properties. The second category is inspired by the Retinex theory. These methods~\cite{retinex_net,kind_plus,kind} usually suffer from a multi-stage training pipeline. They employ different CNNs to decompose the color image, denoise the reflectance, and adjust the illumination, respectively. These CNNs are first trained independently and then connected together to be finetuned end-to-end. The training process is tedious and time-consuming. %While some other Retinex-based deep learning methods~\cite{deep_upe} mainly study corruption-free image enhancment. When applying them in corrupted scenes, the performance drastically degrade.

In addition, these CNN-based methods show limitations in capturing long-range dependencies and non-local self-similarity, which are critical for image restoration. The recently rising deep learning model, Transformer, may provide a possibility to address this drawback of CNN-based methods. However, directly applying original vision Transformers for low-light image enhancement may encounter an issue. The computational complexity is quadratic to the input spatial size. This  computational cost may be unaffordable. %Secondly, if using local window-based Transformer, the receptive fields are limited within position-specific windows. Some content-related tokens may be neglected when computing the self-attention. 
Due to this limitation, some CNN-Transformer hybrid algorithms like SNR-Net~\cite{snr_net} only employ a single global Transformer layer at the lowest spatial resolution of a U-shaped CNN. Thus, the potential of Transformer for low-light image enhancement still remains under-explored. 

To cope with the above problems, we propose a novel method,  Retinexformer, for low-light image enhancement. Firstly, we formulate a simple yet principled One-stage Retinex-based Framework (ORF). We revise the original Retinex model  by introducing perturbation terms to the reflectance and illumination for modeling the corruptions. Our ORF estimates the illumination information and uses it to light up the low-light images. Then ORF employs a corruption restorer to suppress noise, artifacts, under-/over-exposure, and color distortion. Different from  previous Retinex-based deep learning frameworks that suffer from a tedious multi-stage training pipeline, our ORF is trained end-to-end in a one-stage manner. Secondly, we propose an Illumination-Guided Transformer (IGT) to model the long-range dependencies. The key component of IGT is Illumination-Guided Multi-head Self-Attention (IG-MSA). IG-MSA  exploits the illumination representations to direct the computation of self-attention and enhance the interactions between regions of  different exposure levels. Finally, we plug IGT into ORF as the corruption restorer to derive our method, Retinexformer. As shown in Fig.~\ref{fig:teaser}, our Retinexformer surpasses state-of-the-art (SOTA) Retinex-based deep learning methods by large margins on various datasets. Especially on SID~\cite{sid}, SDSD~\cite{sdsd}-indoor, and LOL-v2~\cite{lol_v2}-synthetic, the improvements are over \textbf{6 dB}.

Our contributions can be summarized as follows:
\begin{itemize}
	\vspace{-1mm}
	\item We propose the first Transformer-based algorithm, Retinexformer, for  low-light image enhancement. 
	\vspace{-1mm}
	\item We formulate a one-stage Retinex-based low-light enhancement framework, ORF, that enjoys an easy one-stage training process and models the corruptions well.
	\vspace{-5mm}
	\item We design a new self-attention mechanism, IG-MSA, that utilizes the illumination information as a key clue to guide the modeling of long-range dependences.
	\vspace{-1mm}
	\item Quantitative and qualitative experiments show that our Retinexformer outperforms SOTA methods on thirteen datasets. The results of user study and low-light detection also suggest the practical values of our method.
\end{itemize}
%Traditional Retinex-based methods assume that an color image can be decomposed into two components, \emph{i.e.}, reflectance and illumination. These algorithms treat the reflectance as a plausible approximation of the enhanced image counterpart and then formulat the low-light image enhancement as an illumination estimation problem. However, these methods 

\vspace{-2mm}
\section{Related Work}	
\vspace{-1mm}

\subsection{Low-light Image Enhancement}
\vspace{-1mm}
%The algorithms of low-light image enhancement can be divided into three categories: plain methods, tranditional retinex-based methods, and deep learning-based methods.

\noindent\textbf{Plain Methods.} Plain methods like histogram equalization~\cite{he_1,he_5,he_2,he_3,he} and Gama Correction (GC)~\cite{gc_1,gc_3,gc_2} directly amplify the low visibility and contrast of under-exposed images. Yet, these methods barely consider the illumination factors, making the enhanced images perceptually inconsistent with the real normal-light scenes.

\vspace{0.5mm}
\noindent\textbf{Traditional Cognition Methods.} Different from plain algorithms, conventional  methods~\cite{retinex_1,retinex_4,retinex_3,retinex_5,retinex_6} bear the illumination factors into consideration. They rely on the Retinex theory and treat the reflectance component of the low-light image as a plausible solution of the enhanced result. For example, Guo \emph{et al.}~\cite{lime} propose to refine the initial estimated illumination map by imposing a structure prior on it. %Wang \emph{et al.}~\cite{retinex_6} design an algorithm NPE to preserve the naturalness of details during the enhancement of non-uniform illumination images. 
Yet, these methods naively assume that the low-light images are corruption-free, leading to severe noise and color distortion in the enhancement. Plus, these methods rely on hand-crafted priors, usually requiring careful parameter tweaking and suffering from poor generalization ability.

\begin{figure*}[htp]
	\begin{center}
		\begin{tabular}[t]{c} \hspace{-3.3mm}
			\includegraphics[width=1\textwidth]{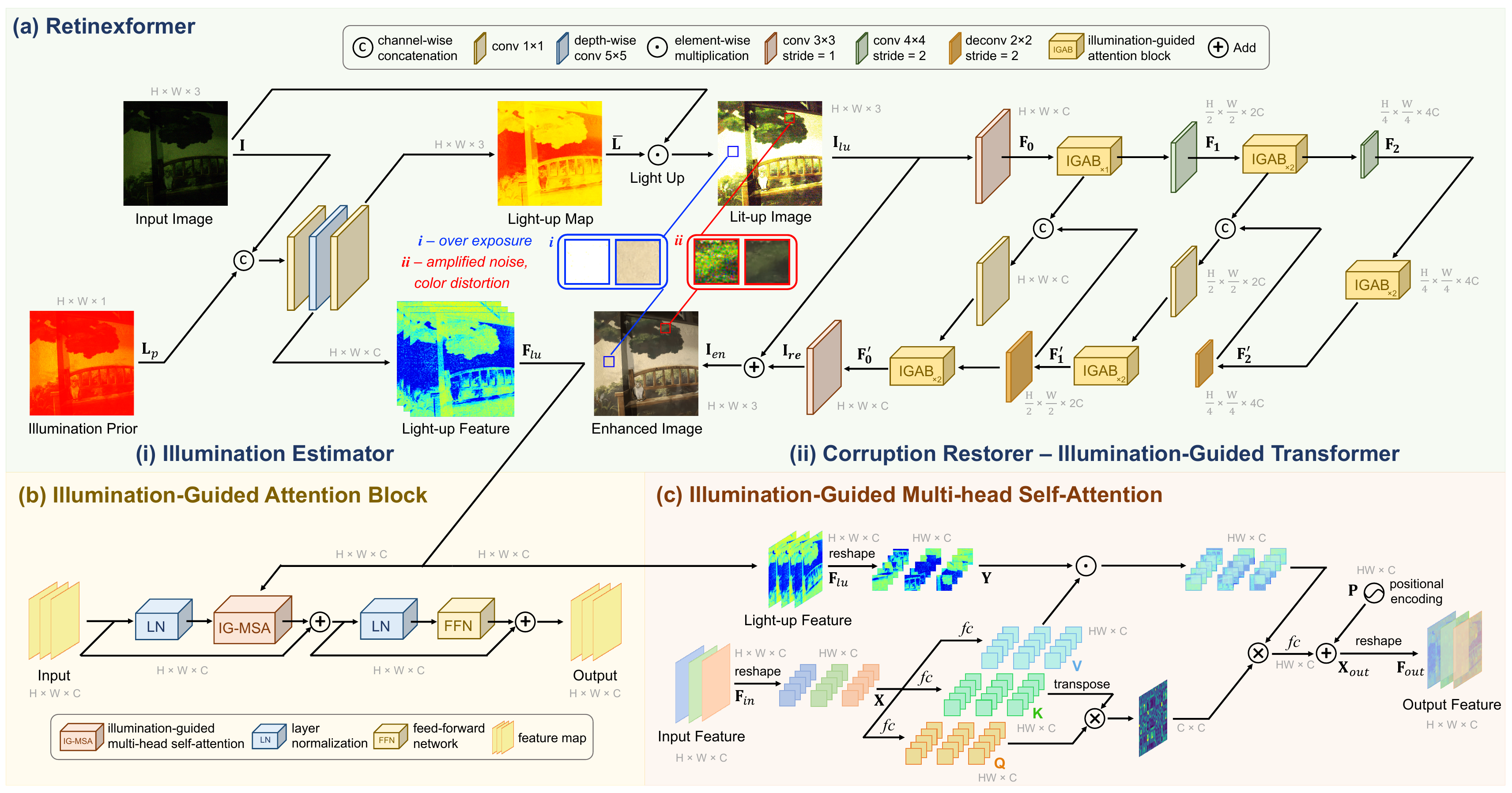}
		\end{tabular}
	\end{center}
	\vspace*{-1mm}
	\caption{\small The overview of our method. (a) Retinexformer adopts the proposed ORF that consists of an illumination estimator (i) and a corruption restorer (ii)  IGT. (b) The basic unit of IGT is IGAB, which is composed of two layer normalization (LN), an IG-MSA and a feed-forward network (FFN). (c) IG-MSA uses the illumination representations captured by ORF  to direct the computation of self-attention.}
	\label{fig:pipeline}
	\vspace{-2mm}
\end{figure*}

\vspace{0.5mm}
\noindent\textbf{Deep Learning Methods.} With the rapid progress of deep learning, CNN~\cite{le-gan,zero_reference,enlightengan,llnet,mbllen,deep_lpf,sharma2021nighttime,deep_upe,mirnet,kind,lednet} has been widely used in low-light image enhancement. For instance, %Lore \emph{et al.}~\cite{llnet} propose a stacked sparse denoising auto-encoder LLNet for joint low-light enhancement and noise suppression. 
Wei \emph{et al.}~\cite{retinex_net} and follow-up works~\cite{kind_plus,kind} combine the Retinex decomposition with deep learning. However, these methods usually suffer from a tedious multi-stage training pipeline. Several CNNs are employed to learn or adjust different components of the Retinex model, respectively. Wang \emph{et al.}~\cite{deep_upe} propose a one-stage Retinex-based  CNN, dubbed DeepUPE, to directly predict the illumination map. Nonetheless, DeepUPE does not consider the corruption factors, leading to amplified noise and color distortion when lighting up  under-exposed photos. In addition, these CNN-based methods also show limitations in capturing long-range dependencies of different  regions.

\vspace{-0.5mm}
\subsection{Vision Transformer}
\vspace{-0.9mm}
The natural language processing model, Transformer, is proposed in ~\cite{vaswani2017attention} for machine translation. In recent years, Transformer and its variants have been applied in many computer vision tasks and achieved impressive results in high-level vision (\emph{e.g.}, image classification~\cite{xcit,arnab2021vivit,global_msa}, semantic segmentation~\cite{cao2021swin,tc_3,SETR}, object detection~\cite{to_1,dy_detr,DETR}, \emph{etc.}) and low-level vision (\emph{e.g.}, image restoration~\cite{dauhst,ipt,restormer}, image synthesis~\cite{gat,transgan,styleswin}, \emph{etc.}). For example, %Wang \emph{et al.}~\cite{uformer} use the basic unit of Swin Transformer~\cite{liu2021swin} to build a U-shaped architecture for image denoising, deblurring, and deraining. 
Xu \emph{et al.}~\cite{snr_net} propose an SNR-aware CNN-Transformer hybrid network, SNR-Net, for low-light image enhancement. However, SNR-Net only employs a single  global Transformer layer at the lowest resolution of a U-shaped CNN due to the enormous computational costs of the vanilla global Transformer. The potential of Transformer has not been fully explored for low-light image enhancement.

\vspace{-1.2mm}
\section{Method}
\vspace{-0.8mm}
Fig.~\ref{fig:pipeline} illustrates the overall architecture of our method. As shown in Fig.~\ref{fig:pipeline} (a), our Retinexformer is based on our formulated One-stage Retinex-based Framework (ORF). ORF consists of an illumination estimator (i) and a corruption restorer (ii). We design an Illumination-Guided Transformer (IGT) to play the role of the corruption restorer. As depicted in Fig.~\ref{fig:pipeline} (b), the basic unit of IGT is Illumination-Guided Attention Block (IGAB), which is composed of two layer normalization (LN), an Illumination-Guided Multi-head Self-Attention (IG-MSA) module, and a feed-forward network (FFN). Fig.~\ref{fig:pipeline} (c) shows the details of IG-MSA.

\vspace{-0.5mm}
\subsection{One-stage Retinex-based Framework}
\vspace{-1mm}
\label{orf}
According to  the Retinex theory. A low-light image $\mathbf{I} \in \mathbb{R}^{H\times W\times 3}$ can be decomposed into a reflectance image  $\mathbf{R} \in \mathbb{R}^{H\times W\times 3}$ and an illumination map  $\mathbf{L} \in \mathbb{R}^{H\times W}$ as
\vspace{-2.2mm}
\begin{equation}
%\begin{aligned}
\small
\mathbf{I} = \mathbf{R} \odot \mathbf{L},
\vspace{-2.3mm}
%\end{aligned}
\label{eq:retinex_ori}
\end{equation}
where $\odot$ denotes the element-wise multiplication. This Retinex model assumes $\mathbf{I}$ is corruption-free, which is inconsistent with the real under-exposed scenes. We analyze that the corruptions mainly steam from two factors. Firstly, the high-ISO and long-exposure imaging settings of dark scenes inevitably introduce noise and artifacts. Secondly, the light-up process may amplify the noise and artifacts and also cause under-/over-exposure  and color distortion, as illustrated in the zoomed-in patch  \textcolor{blue}{$\boldsymbol{i}$} and \textcolor{red}{$\boldsymbol{ii}$} of Fig.~\ref{fig:pipeline} (a).

To model the corruptions, we reformulate Eq.~\eqref{eq:retinex_ori} by introducing a perturbation term for $\mathbf{R}$ and $\mathbf{L}$ respectively, as
\vspace{-2.1mm}
\begin{equation}
\small
\begin{aligned}
\mathbf{I} &= (\mathbf{R} + \mathbf{\hat{R}}) \odot (\mathbf{L} + \mathbf{\hat{L}}) \\
				&= \mathbf{R} \odot \mathbf{L} + \mathbf{R} \odot \mathbf{\hat{L}} + \mathbf{\hat{R}} \odot (\mathbf{L} + \mathbf{\hat{L}}),
\vspace{-2.9mm}
\end{aligned}
\label{eq:retinex_noise}
\end{equation}
where $\mathbf{\hat{R}} \in \mathbb{R}^{H\times W\times 3}$ and $\mathbf{\hat{L}} \in \mathbb{R}^{H\times W}$ denote the perturbations. Similar to \cite{retinex_1,lime,deep_upe}, we regard $\mathbf{R}$ as a well-exposed image. To light up $\mathbf{I}$, we element-wisely multiply the two sides of Eq.~\eqref{eq:retinex_noise} by a light-up map $\mathbf{\bar{L}}$ such that $\mathbf{\bar{L}} \odot \mathbf{{L}} = \mathbf{1}$ as
\vspace{-2mm}
\begin{equation}
\small
%\begin{aligned}
\mathbf{I} \odot \mathbf{\bar{L}} = \mathbf{R} + \mathbf{R} \odot (\mathbf{\hat{L}} \odot \mathbf{\bar{L}}) + (\mathbf{\hat{R}} \odot (\mathbf{L} + \mathbf{\hat{L}})) \odot \mathbf{\bar{L}},
\vspace{-0.5mm}
%\end{aligned}
\label{eq:retinex_light_up}
\end{equation}
where $\mathbf{\hat{R}} \odot (\mathbf{L} + \mathbf{\hat{L}})$ represents the noise and artifacts hidden in the dark scenes and are amplified by $\mathbf{\bar{L}}$. $\mathbf{R} \odot (\mathbf{\hat{L}} \odot \mathbf{\bar{L}})$ indicates the under-/over-exposure and color distortion caused by the light-up process. We simplify Eq.~\eqref{eq:retinex_light_up} as
\vspace{-1mm}
\begin{equation}
%\begin{aligned}
\small
\mathbf{I}_{{lu}}= \mathbf{I} \odot \mathbf{\bar{L}} = \mathbf{R} + \mathbf{C},
\vspace{-1.5mm}
%\end{aligned}
\label{eq:retinex_light_up_2}
\end{equation}
where $\mathbf{I}_{{lu}} \in \mathbb{R}^{H\times W\times 3}$ represents the lit-up image and $\mathbf{C} \in \mathbb{R}^{H\times W\times 3}$ indicates the overall corruption term. Subsequently, we formulate our ORF as
\vspace{-1.2mm}
\begin{equation}
%\begin{aligned}
\small
(\mathbf{I}_{{lu}}, \mathbf{F}_{{lu}}) = \mathcal{E}(\mathbf{I}, \mathbf{L}_{p}), ~~~~\mathbf{I}_{en} = \mathcal{R}(\mathbf{I}_{{lu}}, \mathbf{F}_{{lu}}), 
%\end{aligned}
\vspace{-1.8mm}
\label{eq:orf}
\end{equation}
where $\mathcal{E}$ denotes the illumination estimator and $\mathcal{R}$ represents the corruption restorer. $\mathcal{E}$ takes $\mathbf{I}$ and its  illumination prior map $\mathbf{L}_p \in \mathbb{R}^{H\times W}$ as inputs. $\mathbf{L}_{p}= \text{mean}_c(\mathbf{I})$ where $\text{mean}_c$ indicates  the operation that calculates the mean values for each pixel along the channel dimension. $\mathcal{E}$ outputs the lit-up image $\mathbf{I}_{lu}$ and light-up feature $\mathbf{F}_{lu} \in  \mathbb{R}^{H\times W\times C}$. Then $\mathbf{I}_{lu}$ and $\mathbf{F}_{lu}$ are  fed into $\mathcal{R}$ to restore the corruptions and produce the enhanced image $\mathbf{I}_{en} \in \mathbb{R}^{H\times W\times 3}$.

The architecture of $\mathcal{E}$ is shown in Fig.~\ref{fig:pipeline} (a) (i).  $\mathcal{E}$ firstly uses a $conv$1$\times$1 (convolution with kernel size = 1) to fuse the concatenation of $\mathbf{I}$ and $\mathbf{L}_p$. We notice that the well-exposed regions can provide semantic contextual information for under-exposed regions. Thus, a depth-wise separable $conv$5$\times$5 is adopted to model the interactions of regions with different lighting conditions to generate the light-up feature  $\mathbf{F}_{lu}$.  Then $\mathcal{E}$ uses a $conv$1$\times$1 to aggregate $\mathbf{F}_{lu}$ to produce the light-up map $\mathbf{\bar{L}} \in \mathbb{R}^{H\times W\times 3}$. We set $\mathbf{\bar{L}}$ as a three-channel RGB tensor instead of a single-channel one like ~\cite{retinex_1,lime} to improve its representation capacity in simulating the nonlinearity across RGB channels for color enhancement. Then $\mathbf{\bar{L}}$ is used to light up  $\mathbf{I}$ in Eq.~\eqref{eq:retinex_light_up}. 

%\vspace{1mm}
\vspace{0.5mm}
\noindent\textbf{Discussion}. \textbf{(i)} Different from previous Retinex-based deep learning methods~\cite{ruas,deep_upe,retinex_net,kind_plus,kind}, our ORF estimates $\mathbf{\bar{L}}$ instead of the illumination map $\mathbf{L}$ because if ORF estimates $\mathbf{L}$, then the lit-up image will be obtained by an element-wise division ($\mathbf{I} ./ \mathbf{L}$). Computers are vulnerable to this operation. The values of tensors can be very small (sometimes even equal to 0). The division may easily cause the data overflow issue. Besides, small  errors randomly generated by the computer will be amplified by this operation and lead to inaccurate estimation. Hence, modeling $\mathbf{\bar{L}}$ is more robust.

\noindent\textbf{(ii)} Previous Retinex-based deep learning methods mainly focus on suppressing the corruptions like noise on the reflectance image, \emph{i.e.}, $\mathbf{\hat{R}}$ in Eq.~\eqref{eq:retinex_noise}. They overlook the estimation error on the illumination map, \emph{i.e.}, $\mathbf{\hat{L}}$ in Eq.~\eqref{eq:retinex_noise}, thus easily leading to under-/over-exposure and color distortion during the light up process. In contrast, our ORF considers all these corruptions and employs $\mathcal{R}$ to restore them all.

\vspace{-1mm}
\subsection{Illumination-Guided Transformer}
\vspace{-1mm}
Previous deep learning methods mainly rely on CNNs, showing limitations in capturing long-range dependencies. Some CNN-Transformer hybrid works like SNR-Net~\cite{snr_net} only employ a global Transformer layer at the lowest resolution of a U-shaped CNN due to the enormous computational complexity of global multi-head self-attention (MSA). The potential of Transformer has not been fully explored. To fill this gap, we design an Illumination-Guided Transformer (IGT) to play the role of the corruption restorer $\mathcal{R}$ in Eq.~\eqref{eq:orf}.

\vspace{0.5mm}
\noindent\textbf{Network Structure.} As illustrated in Fig.~\ref{fig:pipeline} (a) (ii), IGT adopts a three-scale U-shaped architecture~\cite{unet}. The input of IGT is the lit-up image $\mathbf{I}_{lu}$. In the downsampling branch, $\mathbf{I}_{lu}$  undergoes a $conv$3$\times$3, an IGAB, a strided $conv$4$\times$4 (for downscaling the features), two IGABs, and a strided $conv$4$\times$4 to generate hierarchical features $\mathbf{F}_{i} \in \mathbb{R}^{\frac{H}{2^i} \times \frac{W}{2^i}  \times 2^{i}C}$ where $i$ = 0, 1, 2. Then $\mathbf{F}_{2}$ passes through two IGABs. Subsequently, a symmetrical structure is designed as the upsampling branch. The $deconv$2$\times$2 with stride = 2 is exploited to upscale the features. Skip connections are used to alleviate the information loss caused by the downsampling branch. The upsampling branch outputs a residual image $\mathbf{I}_{re} \in \mathbb{R}^{H\times W\times 3}$. Then the enhanced image $\mathbf{I}_{en}$ is derived by the sum of $\mathbf{I}_{lu}$ and $\mathbf{I}_{re}$, \emph{i.e.}, $\mathbf{I}_{en}$ = $\mathbf{I}_{lu}$ + $\mathbf{I}_{re}$.

\vspace{1mm}
\noindent\textbf{IG-MSA.} As illustrated in Fig.~\ref{fig:pipeline} (c), the light-up feature $\mathbf{F}_{lu} \in \mathbb{R}^{H\times W\times C}$ estimated by $\mathcal{E}$ is fed into each IG-MSA of IGT. Please note that Fig.~\ref{fig:pipeline} (c) depicts IG-MSA for the largest scale. For smaller scales, $conv$4$\times$4 layers with stride = 2  are used to downscale $\mathbf{F}_{lu}$ to match the spatial size, which is omitted in this figure. As aforementioned, the non-trivial computational cost of global MSA limits the application of Transformer in low-light image enhancement. To tackle this issue, IG-MSA treats a single-channel feature map as a token and then  computes the self-attention.

\begin{figure*}[t]
	\begin{center}
		\begin{tabular}[t]{c} \hspace{-3.5mm}
			\includegraphics[width=1\textwidth]{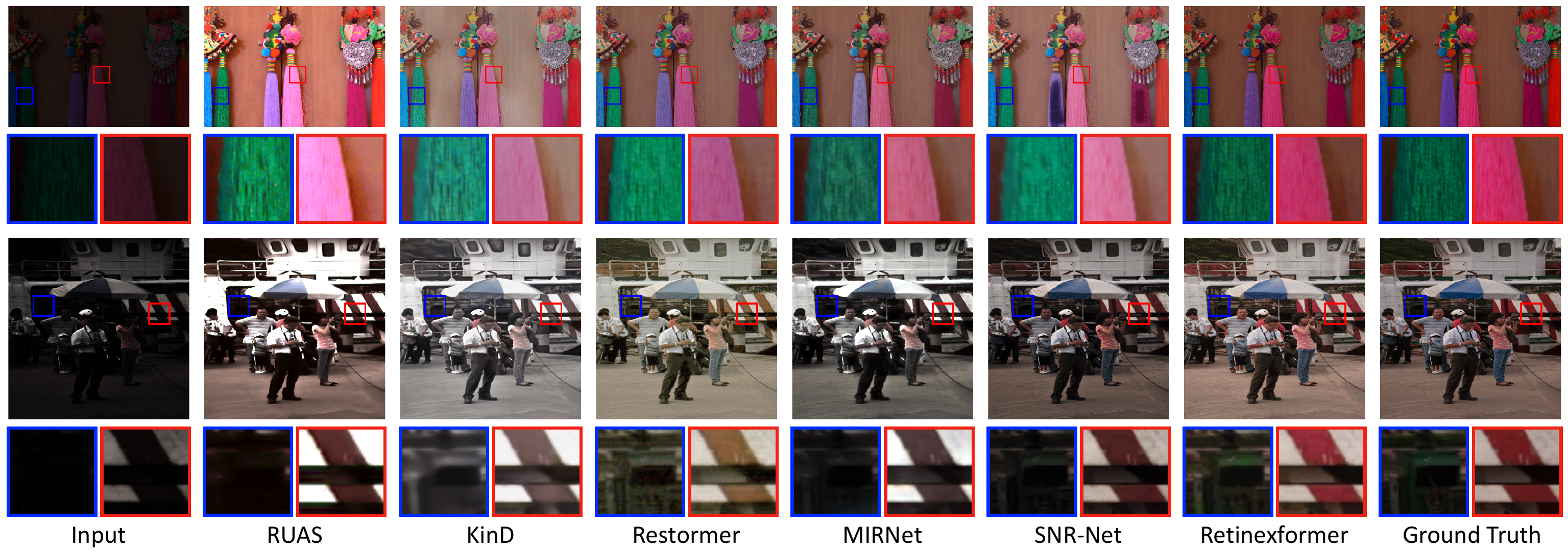}
		\end{tabular}
	\end{center}
	\vspace*{-3mm}
	\caption{\small Results on LOL-v1~\cite{retinex_net} (top) and LOL-v2~\cite{lol_v2} (bottom).   Our method  effectively enhances the visibility and preserves the color.}
	\label{fig:compare}
	\vspace{-1mm}
\end{figure*}

Firstly, the input feature $\mathbf{F}_{in} \in \mathbb{R}^{H\times W\times C}$ is reshaped into tokens $\mathbf{X} \in  \mathbb{R}^{HW\times C}$. Then $\mathbf{X}$ is split into $k$ heads:
\vspace{-0.5mm}
\begin{equation}
\small
%\begin{aligned}
\mathbf{X} = [\mathbf{X}_1,~\mathbf{X}_2,~\cdots,~ \mathbf{X}_k], 
%\end{aligned}
\label{split}
\vspace{-0.5mm}
\end{equation}
where $\mathbf{X}_i \in \mathbb{R}^{HW\times d_k}, d_k = \frac{C}{k},$ and $i = 1, 2, \cdots, k$. Note that Fig.~\ref{fig:pipeline} (c) shows the situation with $k$ = 1 and omits some details for simplification. For each $head_i$, three fully connected ($fc$) layers without $bias$ are used to linearly project $\mathbf{X}_i$ into $query$ elements $\mathbf{Q}_{i} \in \mathbb{R}^{HW \times d_k}$, \emph{key} elements $\mathbf{K}_i \in \mathbb{R}^{HW \times d_k}$, and \emph{value} elements $\mathbf{V}_i \in \mathbb{R}^{HW \times d_k}$ as 
\vspace{-1.5mm}
\begin{equation}
\small
%\begin{aligned}
\mathbf{Q}_i = \mathbf{X}_i\mathbf{W}_{\mathbf{Q}_i}^{\text{T}},~~
\mathbf{K}_i = \mathbf{X}_i\mathbf{W}_{\mathbf{K}_i}^{\text{T}},~~
\mathbf{V}_i = \mathbf{X}_i\mathbf{W}_{\mathbf{V}_i}^{\text{T}},
%\end{aligned}
\label{linear_proj}
\vspace{1.5mm}
\end{equation}
where $\mathbf{W}_{\mathbf{Q}_i}$, $\mathbf{W}_{\mathbf{K}_i}$, and $\mathbf{W}_{\mathbf{V}_i} \in \mathbb{R}^{d_k \times d_k}$ represent the learnable parameters of the $fc$ layers and T denotes the matrix transpose. We notice that different regions of the same image may have different lighting conditions. Dark regions usually have severer corruptions and are more difficult to restore. Regions with better lighting conditions can provide semantic contextual representations to help enhance the dark regions. Thus, we use the light-up feature $\mathbf{F}_{lu}$ encoding illumination information and interactions of regions with different lighting conditions to direct the computation of self-attention. To align with the shape of  $\mathbf{X}$, we also reshape $\mathbf{F}_{lu}$ into $\mathbf{Y} \in \mathbb{R}^{HW\times C}$ and split it into $k$ heads:
\vspace{-0.5mm}
\begin{equation}
\small
%\begin{aligned}
\mathbf{Y} = [\mathbf{Y}_1,~\mathbf{Y}_2,~\cdots,~ \mathbf{Y}_k], 
%\end{aligned}
\label{split_Y}
\vspace{-0.5mm}
\end{equation}
where $\mathbf{Y}_i \in \mathbb{R}^{HW\times d_k}, i = 1, 2, \cdots, k$. Then the self-attention for each $head_i$ is formulated as 
\vspace{-0.5mm}
\begin{equation}
\small
\text{Attention}(\mathbf{Q}_i, \mathbf{K}_i, \mathbf{V}_i, \mathbf{Y}_i) = (\mathbf{Y}_i \odot \mathbf{V}_i)\text{softmax}(\frac{\mathbf{K}_{i}^{\text{T}}\mathbf{Q}_{i}}{\alpha_i}),
\label{attention}
\vspace{-0.5mm}
\end{equation}
where $\alpha_i \in \mathbb{R}^1$ is a learnable parameter that adaptively scales the matrix multiplication. Subsequently, $k$ heads are concatenated to pass through an $fc$ layer and then plus a positional encoding $\mathbf{P} \in \mathbb{R}^{HW\times C}$ (learnable parameters) to produce the output tokens $\mathbf{X}_{out} \in \mathbb{R}^{HW\times C}$. Finally, we reshape $\mathbf{X}_{out}$ to derive the output feature $\mathbf{F}_{out} \in \mathbb{R}^{H\times W\times C}$.

\vspace{1mm}
\noindent\textbf{Complexity Analysis.} We analyze that the computational complexity of our IG-MSA mainly comes from the $k$ computations of the two matrix multiplication in Eq.~\eqref{attention}, \emph{i.e.}, $\mathbb{R}^{d_k\times HW} \times \mathbb{R}^{HW\times d_k}$ and $\mathbb{R}^{HW\times d_k} \times \mathbb{R}^{d_k \times d_k}$. Therefore, the complexity $\mathcal{O}$({IG-MSA}) can be formulated as 
\vspace{-1.2mm}
\begin{equation}
\small
\begin{aligned}
\mathcal{O}(\text{IG-MSA}) &= k\cdot[d_k\cdot(d_k\cdot HW) + HW\cdot(d_k\cdot d_k)], \\
&= 2HWkd_k^2 = 2HWk(\frac{C}{k})^2 = \frac{2HWC^2}{k}.
\end{aligned}
\label{complexity_igmsa}
\vspace{-1mm}
\end{equation}
While the complexity of the global MSA (G-MSA) used by some previous CNN-Transformer methods like SNR-Net is
\vspace{-2.2mm}
\begin{equation}
\small
\mathcal{O}(\text{G-MSA}) = 2(HW)^2C.
\label{complexity_gmsa}
\vspace{-0mm}
\end{equation}
Compare Eq.~\eqref{complexity_igmsa} with Eq.~\eqref{complexity_gmsa}. $\mathcal{O}(\text{G-MSA})$ is quadratic to the input spatial size ($HW$). This burden is expensive and limits the application of Transformer for low-light image enhancement. Therefore, previous CNN-Transformer hybrid algorithms only employ a G-MSA layer at the lowest spatial resolution of a U-shaped CNN to save the computational costs. In contrast, $\mathcal{O}(\text{IG-MSA})$ is linear to the spatial size. This much lower computational complexity enables our IG-MSA to be plugged into each basic unit IGAB of the network. By this means, the potential of Transformer for low-light image enhancement can be further explored.

\begin{table*}[t]
	\centering
	\setlength\tabcolsep{4pt}
	\resizebox{0.995\textwidth}{!}{\hspace{-0.5mm}
		%\begin{tabular}{l|cc|cc|cc|cc|cc|cc|cc|cc}
		\begin{tabular}{l|cc|cc|cc|cc|cc|cc|cc|cc}
			\toprule[0.15em]
			\multirow{2}{*}{Methods}      & \multicolumn{2}{c|}{Complexity}   & \multicolumn{2}{c|}{LOL-v1} & \multicolumn{2}{c|}{LOL-v2-real}  &\multicolumn{2}{c|}{LOL-v2-syn}   &\multicolumn{2}{c|}{SID} &\multicolumn{2}{c|}{SMID} &\multicolumn{2}{c|}{SDSD-in} &\multicolumn{2}{c}{SDSD-out}   \\  & FLOPS (G) & Params (M) & PSNR & SSIM & PSNR & SSIM & PSNR & SSIM & PSNR & SSIM & PSNR & SSIM & PSNR & SSIM & PSNR & SSIM \\ \midrule[0.15em]
			
			SID~\cite{sid}     &13.73 &7.76   & 14.35  &0.436  &13.24  &0.442       &15.04    &0.610 &16.97 &0.591 &24.78 &0.718 &23.29 &0.703 &24.90 &0.693 \\
			
			3DLUT~\cite{3dlut}     &\textcolor{black}{0.075} &\textcolor{black}{0.59}   &14.35  &0.445  &17.59  &0.721   &18.04    &0.800   &20.11 &0.592 &23.86 &0.678 &21.66 &0.655 &21.89 &0.649 \\
			
			DeepUPE~\cite{deep_upe}  & 21.10    &1.02    &14.38  &0.446  &13.27     & 0.452   &15.08 &0.623  & 17.01  &0.604 &23.91 &0.690 &21.70 &0.662 &21.94 &0.698 \\
			
			RF~\cite{rf}     &46.23 &21.54   & 15.23  &0.452  &14.05  &0.458   &15.97    &0.632   &16.44 &0.596 &23.11 &0.681 &20.97 &0.655 &21.21 &0.689 \\
			
			DeepLPF~\cite{deep_lpf}  &5.86 &1.77   &15.28 &0.473  &14.10 &0.480 &16.02 &0.587    &18.07  &0.600 &24.36  &0.688 &22.21 &0.664 &22.76 &0.658\\
			
			IPT~\cite{ipt}    &6887 &115.31   & 16.27        & 0.504        & 19.80        & 0.813        &18.30   &0.811  &20.53 &0.561 &27.03 &0.783 &26.11 &0.831 &27.55 &0.850  \\
			
			UFormer~\cite{uformer}   &12.00 &5.29   & 16.36        & 0.771        & 18.82        & 0.771        & 19.66     &0.871  &18.54 &0.577 &27.20 &0.792 &23.17 &0.859 &23.85 &0.748\\
			
			RetinexNet~\cite{retinex_net}  & 587.47    & 0.84 & 16.77    & 0.560   & 15.47        & 0.567  &17.13  &0.798  &16.48 &0.578 &22.83 &0.684 &20.84 &0.617 &20.96 &0.629 \\
			
			Sparse~\cite{lol_v2}     &53.26 &2.33   &17.20  &0.640  &20.06  &0.816   &22.05    &0.905     &18.68 &0.606 &25.48 &0.766 &23.25 &0.863 &25.28 &0.804 \\
			
			EnGAN~\cite{enlightengan}  &61.01  &114.35  &17.48  &0.650    &18.23  &0.617   &16.57   &0.734  &17.23 &0.543  &22.62 &0.674 &20.02 &0.604 &20.10 &0.616\\
			
			RUAS~\cite{ruas}     &\textcolor{black}{0.83} &\textcolor{black}{0.003}   &18.23  &0.720  &18.37  &0.723   &16.55    &0.652    &18.44 &0.581 &25.88 &0.744 &23.17 &0.696 &23.84 &0.743 \\

			%HDRNet~\cite{hdrnet}     &- &-   &-  &-  &- &-   &-   &-    &- &- &- &- &- &- &- &- \\
			FIDE~\cite{fide}     &28.51 &8.62   & 18.27  &0.665  &16.85  &0.678   &15.20    &0.612    &18.34 &0.578 &24.42 &0.692 &22.41 &0.659 &22.20 &0.629 \\
			
			DRBN~\cite{drbn} &48.61  &5.27   & 20.13   & 0.830    &20.29   & 0.831    & 23.22     & 0.927   &19.02 &0.577  &26.60 &0.781  &24.08  &0.868 &25.77 &0.841\\
			
			KinD~\cite{kind} &34.99    &8.02   &20.86 &0.790  &14.74 &0.641  &13.29 &0.578  &18.02 &0.583 &22.18 &0.634 &21.95 &0.672 &21.97 &0.654 \\

			Restormer~\cite{restormer}   &144.25 &26.13   &22.43        &0.823        &19.94         &0.827         &21.41      &0.830         &22.27 &0.649 &26.97 &0.758 &25.67 &0.827 &24.79 &0.802 \\
			
			MIRNet~\cite{mirnet}   &785 &31.76    &24.14   &0.830  &20.02   &0.820  &21.94  &0.876   &20.84 &0.605   &25.66 &0.762 &24.38  &0.864 &27.13 &0.837 \\    
			SNR-Net~\cite{snr_net}   &26.35   &4.01  &\textcolor{blue}{24.61} &\textcolor{blue}{0.842}  &\textcolor{blue}{21.48}  &\textcolor{red}{0.849}  &\textcolor{blue}{24.14} &\textcolor{blue}{0.928}  &\textcolor{blue}{22.87} &\textcolor{blue}{0.625} &\textcolor{blue}{28.49} &\textcolor{blue}{0.805} &\textcolor{blue}{29.44} &\textcolor{blue}{0.894} &\textcolor{blue}{28.66} &\textcolor{blue}{0.866} \\ \midrule[0.15em]
			\textbf{Retinexformer}      &15.57  &1.61   &\textcolor{red}{25.16}      &\textcolor{red}{0.845}        &\textcolor{red}{22.80}      &\textcolor{blue}{0.840}    &\textcolor{red}{25.67} &\textcolor{red}{0.930} &\textcolor{red}{24.44} &\textcolor{red}{0.680} &\textcolor{red}{29.15} &\textcolor{red}{0.815}  &\textcolor{red}{29.77} &\textcolor{red}{0.896} &\textcolor{red}{29.84} &\textcolor{red}{0.877} \\ \bottomrule[0.15em]
	\end{tabular}}
	\vspace{2mm}
	\caption{Quantitative comparisons on LOL (v1~\cite{retinex_net} and v2~\cite{lol_v2}), SID~\cite{sid}, SMID~\cite{smid}, and SDSD~\cite{sdsd} (indoor and outdoor) datasets. The highest result is in \textcolor{red}{red} color while the second highest result is in \textcolor{blue}{blue} color. Our Retinexformer significantly outperforms SOTA algorithms. }\label{tab:quantitative}
	\vspace{-3mm}
\end{table*}

\section{Experiment}
\vspace{-1mm}
\subsection{Datasets and Implementation Details}
\vspace{-1mm}
We eveluate our method on LOL (v1~\cite{retinex_net} and v2~\cite{lol_v2}), SID~\cite{sid}, SMID~\cite{smid},  SDSD~\cite{sdsd}, and FiveK~\cite{fivek} datasets.

\begin{figure*}[t]
	\begin{center}
		\begin{tabular}[t]{c} \hspace{-3.3mm}
			\includegraphics[width=1.0\textwidth]{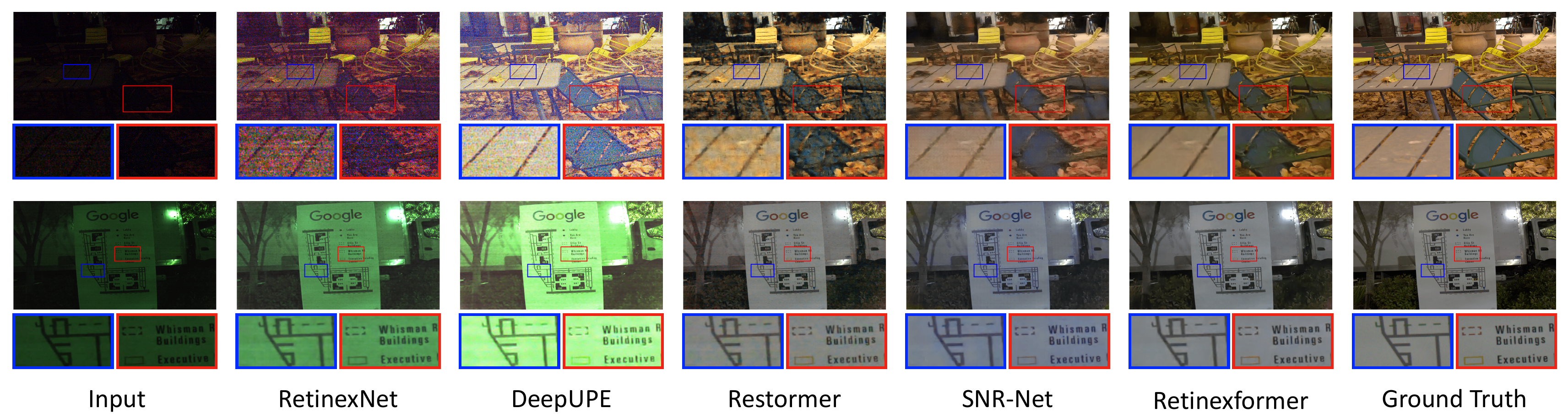}
		\end{tabular}
	\end{center}
	\vspace*{-4mm}
	\caption{\small Visual results on SID~\cite{sid} (top) and SMID~\cite{smid} (bottom). Previous methods either collapse by noise, or distort color, or produce blurry and under-/over-exposed images. While our algorithm can effectively remove the noise and reconstruct well-exposed image details. }
	\label{fig:compare_sid_smid}
	\vspace{-3mm}
\end{figure*}

\noindent{\textbf{LOL.}} The LOL dataset has v1 and v2 versions. LOL-v2 is divided into real and synthetic subsets. The training and testing sets are split in proportion to 485:15, 689:100,
and 900:100 on LOL-v1, LOL-v2-real, and LOL-v2-synthetic. %The train and test set of LOL-v2-real include 689 and 100 image pairs. %Most low-light images are captured from a variety of scenes by changing the ISO and exposure time, while other parameters are fixed. LOL-v2-synthetic is created by analyzing the illumination distribution of low-light images and then synthesizing from RAW images.

%\noindent{\textbf{LOL.}} The LOL dataset has two versions, v1~\cite{retinex_net} and v2~\cite{lol_v2}. We adopt the v2 version to evaluate our method for two reasons. Firstly, the v1 version is a subset of the v2 version. Secondly, the v2 version is larger and more diverse. It can better evaluate the performance and generalization ability of algorithms. The LOL dataset is divided into LOL-real and LOL-synthetic. Most low-light images of LOL-real are  captured from a variety of scenes by changing the ISO and exposure time, while other parameters are fixed. LOL-real provides 689 low-/normal-light image pairs for training and 100 images for testing. LOL-synthetic is obtained by analyzing the illumination distribution of low-light images and then synthesizing from RAW images. LOL-synthetic includes 900 images for training and 100 images for testing.

%\vspace{0.5mm}
\noindent{\textbf{SID.}} The subset of SID dataset captured by Sony $\alpha$7S II  camera is adopted for evaluation. There are 2697 short-/long-exposure RAW image pairs. The low-/normal-light RGB images are obtained by using the same in-camera signal processing of SID~\cite{sid} to transfer RAW to RGB. 2099 adn 598 image pairs are used for training and testing.

%\vspace{0.5mm}
\noindent{\textbf{SMID.}} The SMID benchmark collects 20809 short-/long-exposure RAW image pairs. We also transfer the RAW data to low-/normal-light RGB image pairs. 15763 pairs are used for training and the left pairs are adopted for testing.

%\vspace{0.5mm}
\noindent{\textbf{SDSD.}} We adopt the static version of SDSD. It is captured by a Canon EOS 6D Mark II camera with an ND filter. SDSD contains indoor and outdoor subsets. We respectively use 62:6 and 116:10 low-/normal-light video pairs for training and testing on SDSD-indoor and SDSD-outdoor.

\noindent{\textbf{FiveK.}} MIT-Adobe FiveK dataset is divided into training and testing sets with 4500 and 500 low-/normal-light image pairs. These images are manually adjusted by five photographers (labelled as A$\sim$E). We use experts
C’s adjusted images as reference and adopt the sRGB output mode.

In addition to the above eight benchmarks, we test our method on five datasets: LIME~\cite{lime}, NPE~\cite{retinex_6}, MEF~\cite{mef}, DICM~\cite{he_4}, and VV~\cite{vv} that have no ground truth.

\noindent{\textbf{Implementation Details.}} We implement Retinexformer by PyTorch~\cite{pytorch}. The model is trained with the Adam~\cite{adam} optimizer ($\beta_1$ = 0.9 and $\beta_2$ = 0.999) for 2.5 $\times$ 10$^{5}$ iterations. The learning rate is initially set to 2$\times 10^{-4}$ and then steadily decreased to 1$\times 10^{-6}$ by the cosine annealing scheme~\cite{sgdr} during the training process. Patches at the size of 128$\times$128 are randomly cropped from the low-/normal-light image pairs as training samples. The batch size is 8. The training data is augmented with random rotation and flipping. The training objective is to minimize the mean absolute error (MAE) between the enhanced image and ground truth. We adopt the peak signal-to-noise ratio (PSNR) and structural similarity (SSIM)~\cite{ssim} as the evaluation metrics.

\begin{table}[h]
	\begin{center}
		\vspace{1.5mm}
		\resizebox{0.47\textwidth}{!}{\noindent
			\begin{tabular}{l | c c c c c}
				\toprule
				Methods  &DeepUPE~\cite{deep_upe} &MIRNet~\cite{mirnet} &SNR-Net~\cite{snr_net} &Restormer~\cite{restormer} &\bf Ours \\
				\midrule
				PSNR (dB) &23.04 &23.73  &23.81 &24.13 &\bf 24.94 \\
				FLOPS (G) &21.10 &785.0  &26.35 &144.3  &\bf 15.57 \\
				\bottomrule
		\end{tabular}}
		\vspace{2.5mm}
		\caption{\small Results on the FiveK~\cite{fivek} dataset with sRGB output mode.}
		\label{tab:adobe_5k}
	\end{center}\vspace{-8.5mm}
\end{table}

%将四个表格合并成一个
\begin{table*}[t]\hspace{-2.2mm}
	% subfloat b - mask representation
	\subfloat[\small User study scores on seven benchmarks.\label{user_study}]{\vspace{1mm}
		\scalebox{0.59}{
			\tabcolsep=1.3mm
			\begin{tabular}{lcccccccc}
				\toprule[0.15em]
				\rowcolor{color3} {Methods}      &L-v1 &L-v2-R  &L-v2-S  &SID  &SMID  &SD-in  &SD-out  &Mean\\
				\midrule[0.15em]
				
				EnGAN~\cite{enlightengan}  		&2.43 	&1.39   &2.13  &1.04  &2.78  &1.83   &1.87 &1.92  \\
				
				RetinexNet~\cite{retinex_net}    	&2.17 	&1.91   &1.13  &1.09 &2.35  &\textcolor{red}{3.96}      &\textcolor{blue}{3.74}   & 2.34\\

				DRBN~\cite{drbn}    		&2.70 	&2.26  &\textcolor{blue}{3.65}  &1.96  &2.22  &2.78   &2.91  &2.64 \\
				
				FIDE~\cite{fide}    			&2.87 	&2.52   &3.48  &2.22 &2.57  &3.04       &2.96    & 2.81\\
				
				KinD~\cite{kind}  &2.65 &2.48  &3.17 &1.87  &3.04  &3.43   &3.39   &2.86 \\
				
				MIRNet~\cite{mirnet}		&2.96 	&3.57   &3.61  &2.35  &2.09  &2.91       &3.09    &2.94 \\
				
				Restormer~\cite{restormer}	&3.04 	&3.48   &3.39  &2.43  &3.17  &2.48   &2.70 &2.96 \\
				
				RUAS~\cite{ruas}	&\textcolor{red}{3.83}   &3.22    &2.74  &2.26  &\textcolor{blue}{3.48}  &3.39  &3.04 &3.14  \\
				
				SNR-Net~\cite{snr_net}	&3.13 	&\textcolor{blue}{3.83}   &3.57 &\textcolor{blue}{3.04}  &3.30 &2.74 &3.17 &\textcolor{blue}{3.25}  \\
				
				%GT &3.91 &4.09   &63.6 &85.3  &77.5 &59.1 &54.1 &59.6  \\
				
				\midrule[0.15em]
				\textbf{Retinexformer}   &\textcolor{blue}{3.61} &\textcolor{red}{4.17} &\textcolor{red}{3.78}     &\textcolor{red}{3.39}       &\textcolor{red}{3.87}     &\textcolor{blue}{3.65}  &\textcolor{red}{3.91}  &\textcolor{red}{3.77} \\ 
				\bottomrule[0.15em]
	\end{tabular}}}\vspace{0mm}\hspace{2mm}
	% subfloat a - RoIAlign (ResNet-50-C5)
	\subfloat[\small Low-light detection results on ExDark~\cite{exdark} enhanced by different algorithms. \label{exdark}]{\vspace{1mm}
		\scalebox{0.64}{%\vspace{2mm}
			\tabcolsep=1.3mm
			\begin{tabular}{lccccccccccccc}
				\toprule[0.15em]
				\rowcolor{color3} {Methods}      &Bicycle &Boat  &Bottle  &Bus  &Car  &Cat  &Chair  &Cup  &Dog &Motor &People &Table & Mean\\
				\midrule[0.15em]
				
				%Baseline    &74.1 &63.4   &63.0  &83.5  &76.4  &55.9   &51.8  &59.1 &\textcolor{blue}{66.3} &61.1 &68.4 &43.8 &63.9  \\
				
				MIRNet~\cite{mirnet}     &71.8 &63.8   &62.9  &81.4  &71.1  &58.8       &58.9    &61.3 &63.1 &52.0 &68.8 &45.5 &63.6  \\
				
				RetinexNet~\cite{retinex_net}     &73.8 &62.8   &64.8  &84.9  &\textcolor{red}{80.8}  &53.4       &57.2    &\textcolor{red}{68.3} &61.5 &51.3 &65.9 &43.1 &64.0  \\
				
				RUAS~\cite{ruas}  & 72.0    &62.2    &65.2  &72.9  &\textcolor{blue}{78.1}  &57.3   &\textcolor{red}{62.4} &\textcolor{blue}{61.8}  &60.2  &61.5 &69.4 &\textcolor{blue}{46.8} &64.2  \\
				
				Restormer~\cite{restormer}     &\textcolor{blue}{76.2} &65.1   &64.2  &84.0  &76.3  &59.2   &53.0    &58.7   &66.1 &62.9 &68.6 &45.0 &64.9  \\
				
				KinD~\cite{kind}     &72.2 &\textcolor{blue}{66.5}   &58.9  &83.7  &74.5  &55.4   &\textcolor{blue}{61.7}   &61.3   &63.8 &63.0 &\textcolor{red}{70.5} &\textcolor{red}{47.8} &65.0  \\

				ZeroDCE~\cite{zero_reference}     &75.8 &\textcolor{blue}{66.5}   &\textcolor{blue}{65.6} &84.9  &77.2 &56.3 &53.8 &59.0    &63.5  &\textcolor{blue}{64.0} &68.3  &46.3 &65.1 \\
				
				SNR-Net~\cite{snr_net}  &75.3 &64.4   &63.6 &\textcolor{blue}{85.3}  &77.5 &59.1 &54.1 &59.6    &\textcolor{blue}{66.3}  &\textcolor{red}{65.2} &69.1  &44.6 &65.3 \\
				
				SCI~\cite{sci}    &74.6 &65.3   &65.8        & \textcolor{red}{85.4}        & 76.3        &\textcolor{blue}{59.4}        &57.1   &60.5  &65.6 &63.9 &69.1 &45.9 &\textcolor{blue}{65.6}   \\
				\midrule[0.15em]
				\textbf{Retinexformer}      &\textcolor{red}{76.3}  &\textcolor{red}{66.7}  &\textcolor{red}{65.9}      &84.7        & 77.6     &\textcolor{red}{61.2}   &53.5 &60.7 &\textcolor{red}{67.5} & 63.4 &\textcolor{blue}{69.5} &46.0  &\textcolor{red}{66.1} \\ 
				\bottomrule[0.15em]
		\end{tabular}}}\vspace{0mm}
	\vspace{-2.5mm}
	\caption{\small (a) compares the human perception quality of various low-light  enhancement algorithms. (b) compares the preprocessing effects of different methods on high-level vision understanding. The highest results are in \textcolor{red}{red} color and the second highest results are in \textcolor{blue}{blue} color.}
	\label{tab:us_lod}\vspace{-2mm}
\end{table*}

\begin{figure*}[t]
	\begin{center}
		\begin{tabular}[t]{c} \hspace{-4mm}
			\includegraphics[width=1\textwidth]{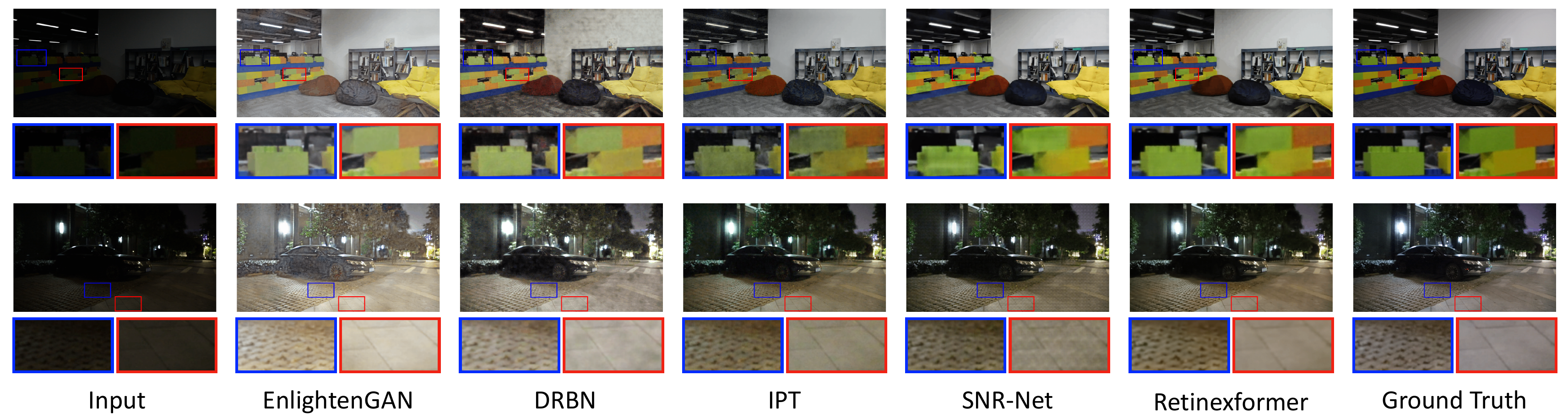}
		\end{tabular}
	\end{center}
	\vspace*{-4mm}
	\caption{\small Visual resulst on SDSD~\cite{sdsd}-indoor (top) and out-door (bottom). Other algorithms either generate over-exposed and  noisy images, or introduce black spot corruptions and unnatural artifacts. While Retinexformer can restore well-exposed structural contents and textures.}
	\label{fig:compare_sdsd}
	\vspace{-4mm}
\end{figure*}

\begin{figure}[h]
	\begin{center}
		\begin{tabular}[t]{c} \hspace{-3.4mm} 
			\includegraphics[width=0.48\textwidth]{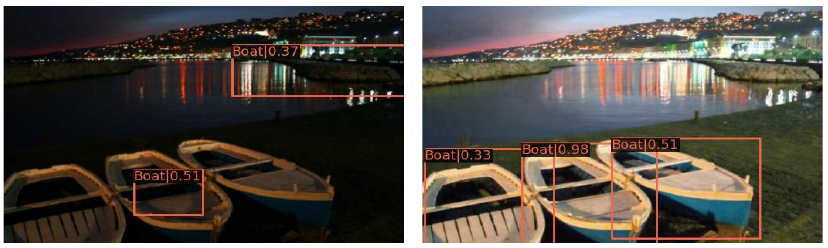}
		\end{tabular}
	\end{center}
	\vspace{-2mm}
	\caption{\small Visual comparison of object detection in low-light (left) and enhanced (right) scenes by our method on the Exdark dataset. }
	\label{fig:det}
	\vspace{-3mm}
\end{figure}

%将四个表格合并成一个
\begin{table*}[t]\hspace{-2.5mm}
	% Table  a - break down ablation
	\subfloat[\small Break-down ablation to higher performance. \label{tab:breakdown}]{\vspace{1mm}
		\scalebox{0.57}{
			\begin{tabular}{c c c  c c c c}
				%\small
				\toprule
				\rowcolor{color3} Baseline-1 &ORF &IG-MSA   &PSNR &SSIM &Params (M) &FLOPS (G) \\
				\midrule
				\checkmark & & &26.47 &0.843 &1.01 &9.18 \\
				\checkmark  &\checkmark & &27.92 &0.857 &1.27 &11.37 \\
				\checkmark & &\checkmark &28.86 &0.868 &1.34 &13.38 \\
				\checkmark  &\checkmark &\checkmark &\bf 29.84 &\bf 0.877 &1.61 &15.57\\
				\bottomrule
	\end{tabular}}}\hspace{1mm}\vspace{0mm}
	% Table b - study of ORF
	\subfloat[\small Ablation of the proposed ORF.\label{tab:orf}]{\vspace{1mm}
	\scalebox{0.57}{
		\begin{tabular}{l c c c c}
			\toprule
			\rowcolor{color3} Method &~$\mathbf{I}_{lu} = \mathbf{I}$~ &~$\mathbf{I}_{lu} = \mathbf{I} ./ \mathbf{L}$~ &~$\mathbf{I}_{lu} = \mathbf{I} \odot \mathbf{\bar{L}}$~ &~$+ \mathbf{F}_{lu}$~\\
			\midrule
			PSNR &28.86 &28.97 &29.26 &\bf 29.84  \\
			SSIM &0.868 &0.868 &0.870 &\bf 0.877  \\
			Params (M)  &1.34 &1.61 &1.61 &1.61 \\
			FLOPS (G)  &13.38 &14.01 &14.01 &15.57 \\
			\bottomrule
		\end{tabular}}}\hspace{1mm}\vspace{0mm}
	\iffalse
	% Table c - study of MSA
		\subfloat[\small Ablation of self-attention schemes.\label{tab:attention}]{
		\vspace{1mm}\scalebox{0.59}{
			\begin{tabular}{l c c c}
				\toprule
				\rowcolor{color3} Method &Baseline-2 &~~G-MSA~~ &~~IG-MSA~~\\
				\midrule
				PSNR &27.92 &32.67 &\bf 29.84  \\
				SSIM &0.857 &0.912 &\bf 0.877  \\
				Params (M)  &1.27 &1.61 &1.61  \\
				FLOPS (G)  &11.37 &14.99 &15.57 \\
				\bottomrule
	\end{tabular}}}\vspace{0mm}
	\fi
	% Table c - study of MSA
	\subfloat[\small Ablation of self-attention schemes.\label{tab:msa}]{
		\vspace{1mm}\scalebox{0.57}{
			\begin{tabular}{l c c c c}
				\toprule
				\rowcolor{color3} Method &Baseline-2 &G-MSA &W-MSA &IG-MSA\\
				\midrule
				PSNR &27.92 &28.43  &28.65 &\bf 29.84  \\
				SSIM &0.857 &0.841 &0.845 &\bf 0.877  \\
				Params (M)  &1.27 &1.61  &1.61 &1.61  \\
				FLOPS (G)  &11.37 &17.65 &16.43 &15.57 \\
				\bottomrule
	\end{tabular}}}\vspace{0mm}
	\vspace{1mm}
	\caption{\small We conduct ablation study  on the  SDSD~\cite{sdsd}-outdoor dataset. PSNR, SSIM, Params, and FLOPS (size = 256$\times$256) are reported.}
	\label{tab:ablations}\vspace{-2mm}
\end{table*}

\iffalse
\begin{figure*}[t]
	\begin{center}
		\begin{tabular}[t]{c} \hspace{-3mm}
			\includegraphics[width=1.0\textwidth]{img/det_compare.pdf}
		\end{tabular}
	\end{center}
	\vspace*{-4.5mm}
	\caption{\small Visual comparisons of enhancement-based low-light object detection on the  ExDark~\cite{exdark} dataset. Different enhancement methods serve as the preprocessing modules to enhance the low-light image before detection. Please zoom in for better visualization performance.}
	\label{fig:compare_det}
	\vspace{-2mm}
\end{figure*}
\fi

\begin{figure*}[t]
	\begin{center}
		\begin{tabular}[t]{c} \hspace{-3.5mm}
			\includegraphics[width=0.99\textwidth]{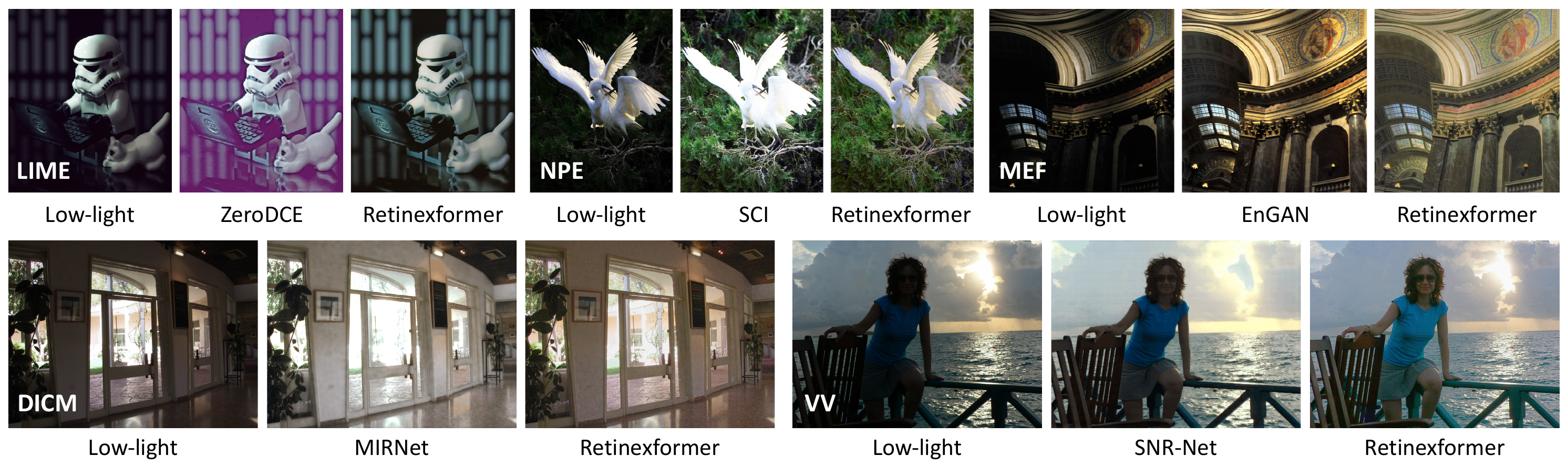}
		\end{tabular}
	\end{center}
	\vspace*{-3mm}
	\caption{\small Visual results on the LIME~\cite{lime}, NPE~\cite{retinex_6}, MEF~\cite{mef}, DICM~\cite{he_4}, and VV~\cite{vv} datasets. Our Retinexformer performs better.}
	\label{fig:no_gt}
	\vspace{-4mm}
\end{figure*}

%\vspace{-1.8mm}
\subsection{Low-light Image Enhancement}
%\vspace{-1.2mm}
\noindent\textbf{Quantitative Results.} We quantitatively compare the proposed method with a wide range of SOTA enhancement algorithms in Tab.~\ref{tab:quantitative} and Tab.~\ref{tab:adobe_5k}. Our Retinexformer significantly outperforms SOTA methods on eight datasets while requiring moderate computational and memory costs. 

When compared with the recent best method SNR-Net, our method achieves 0.55, 1.32, 1.53, 1.57, 0.66, 0.33, 1.18, and 1.13 dB improvements on LOL-v1, LOL-v2-real, LOL-v2-synthetic, SID, SMID, SDSD-indoor, SDSD-outdoor, and FiveK datasets. However, our method only costs 40\% (1.61 / 4.01) Parmas and 59\% (15.57 / 26.35) FLOPS.

When compared with SOTA Retinex-based deep learning methods (including DeepUPE~\cite{deep_upe}, RetinexNet~\cite{retinex_net}, RUAS~\cite{ruas}, and KinD~\cite{kind}), our Retinexformer yields 4.30, 4.43, 8.54, 6.00, 3.27, 6.60, and 6.00 dB improvements on the seven benchmarks in Tab.~\ref{tab:quantitative}. Especially on SID and SDSD datasets that are severely corrupted by noise and artifacts, the improvements are \textbf{over 6 dB}, as plotted in Fig.~\ref{fig:teaser}. 

When compared with SOTA Transformer-based image restoration algorithms (including IPT~\cite{ipt}, Uformer~\cite{uformer}, and Restormer~\cite{restormer}), our Retinexformer gains by 2.73, 2.86, 4.26, 2.17, 1.95, 3.66, and 2.29 dB on the seven datasets in Tab.~\ref{tab:quantitative}. Yet, Retinexformer only requires  1.4\% and 6.2\% Params, 0.2\% and 10.9\% FLOPS of IPT and Restormer.

All these results clearly suggest the outstanding effectiveness and efficiency advantage of our Retinexformer.

\vspace{1mm}
\noindent\textbf{Qualitative Results.} The visual comparisons of Retinexformer and SOTA algorithms are shown in Fig.~\ref{fig:compare},~\ref{fig:compare_sid_smid},~\ref{fig:compare_sdsd}, and \ref{fig:no_gt}. Please zoom in for a better view. Previous methods either cause color distortion like RUAS in Fig.~\ref{fig:compare}, or contain over-/under-exposed regions and fail to suppress the noise like RetinexNet and DeepUPE in Fig.~\ref{fig:compare_sid_smid}, or generate blurry images like Restormer and SNR-Net in Fig.~\ref{fig:compare_sid_smid}, or introduce black spots and unnatural artifacts like DRBN~\cite{drbn} and SNR-Net in Fig.~\ref{fig:compare_sdsd}. In contrast, our Retinexformer can effectively enhance the poor visibility and low contrast or low-light regions, reliably remove the noise without introducing  spots and artifacts, and robustly preserve the color.

Please note that the five datasets in Fig.~\ref{fig:no_gt} have no ground truth. Therefore, the visual results in Fig.~\ref{fig:no_gt} are more convincing and fair to justify the effectiveness. As can be seen that our method performs better than other SOTA supervised and unsupervised algorithms across various scenes.

\vspace{1mm}
\noindent\textbf{User Study Score.} We conduct a user study to quantify the human subjective visual perception quality of the enhanced low-light images from the seven datasets. 23 human subjects are invited to score the visual quality of the enhanced results, independently. These testers are told to observe the results from: \textbf{(i)} whether the results contain under-/over-exposed regions, \textbf{(ii)} whether the results contain color distortion, and \textbf{(iii)} whether the results are corrupted by noise or artifacts. The scores range from 1 (worst) to 5 (best). For each low-light image, we display it and the results enhanced by various algorithms but without their names to the human testers. There are 156 testing images in total. The user study scores are reported in Tab.~\ref{user_study}. Our Retinexformer achieves the highest score on average. Besides, our results are  most favored by the human subjects on LOL-v2-real (L-v2-R), LOL-v2-synthetic (L-v2-S), SID, SMID, and SDSD-outdoor (SD-out) datasets and second most favored on LOL-v1 (L-v1) and SDSD-indoor (SD-in) benchmarks.

\vspace{-0.5mm}
\subsection{Low-light Object Detection}
\vspace{-0.5mm}
\noindent\textbf{Experiment Settings.} We conduct low-light object detection experiments on the ExDark~\cite{exdark} dataset to compare the preprocessing  effects of different enhancement algorithms for high-level vision understanding. The ExDark dataset consists of 7363 under-exposed images annotated with 12 object category bounding boxes. 5890 images are selected for training while the left 1473 images are used for testing. YOLO-v3~\cite{yolov3} is employed as the detector and trained from scratch. Different low-light enhancement methods serve as the preprocessing modules with fixed parameters. %Pre-trained weights are not used for the detector.%Since we focus on studying the effect of enhancement methods, pre-trained weights for the detector are not used. Instead, all detectors are trained from scratch. 

\vspace{0.5mm}
\noindent\textbf{Quantitative Results.} The average precision (AP) scores are listed  in Tab.~\ref{exdark}. Our Retinexformer achieves the highest  result on average, 66.1 AP, which is 0.5 AP higher than the recent best self-supervised method SCI~\cite{sci} and  0.8 AP higher than the recent best fully-supervised method SNR-Net~\cite{snr_net}. Besides, Retinexformer yields the best results on five object categories: bicycle, boat, bottle, cat, and dog.  % and also reveal the practical values of our Retinexformer.

\vspace{0.5mm}
\noindent\textbf{Qualitative Results.} Fig.~\ref{fig:det} depicts a visual comparison of detection results in the low-light (left) scene and the scene enhanced (left) by  Retinexformer. The detector easily misses some boats or predicts inaccurate locations on the under-exposed image. In contrast, the detector can reliably predict well-placed  bounding boxes to cover all boats on the image enhanced by our Retinexformer, showing the effectiveness of our  method in benefiting high-level vision.

 %These results verify that our Retinexformer can generate more visually pleasing results.

\vspace{-0.2mm}
\subsection{Ablation Study}
\vspace{-0.3mm}
We conduct ablation study on the SDSD-outdoor dataset for the good convergence and stable performance of Retinexformer on it. The results are reported in Tab.~\ref{tab:ablations}.

\vspace{0.5mm}
\noindent\textbf{Break-down Ablation.} We conduct a break-down ablation to study the effect of each component towards higher performance, as shown in Tab.~\ref{tab:breakdown}.  Baseline-1 is derived by removing ORF and IG-MSA from  Retinexformer. When we respectively apply ORF and IG-MSA, baseline-1 achieves 1.45 and 2.39 dB improvements. When jointly exploiting the two techniques, baseline-1 gains by 3.37 dB. This evidence suggests the effectiveness of our ORF and IG-MSA.

\vspace{0.5mm}
\noindent\textbf{One-stage Retinex-based Framework.} We conduct an ablation to study ORF. The results are listed in Tab.~\ref{tab:orf}. We first remove ORF from Retinexformer and set the input of $\mathcal{R}$ as $\mathbf{I}_{lu} = \mathbf{I}$. The model yields 28.86 dB. Then we apply ORF but set $\mathcal{E}$ to estimate the illumination map $\mathbf{L}$. The input of $\mathcal{R}$ is $\mathbf{I}./\mathbf{L}$ where $./$ indicates the element-wise division. To avoid exceptions thrown by computer, we add $\mathbf{L}$ with a small constant $\epsilon$ = 1$\times$10$^{-4}$. Yet, as analyzed in Sec.~\ref{orf}, the computer is vulnerable to the division of small values. Thus, the model obtains a limited improvement of 0.11 dB. To tackle this issue, we estimate the light-up map $\mathbf{\bar{L}}$ and set the input of $\mathcal{R}$ as $\mathbf{I}_{lu}$ = $\mathbf{I}\odot \mathbf{\bar{L}}$. The model gains by 0.40 dB. After using  $\mathbf{F}_{lu}$ to direct $\mathcal{R}$, the model continues to achieve an improvement of 0.58 dB in PSNR and 0.007 in SSIM.

\vspace{0.5mm}
\noindent\textbf{Self-Attention Scheme.} We conduct an ablation to study the effect of the self-attention scheme. The results are reported in Tab.~\ref{tab:msa}. Baseline-2 is obtained by removing IG-MSA from Retinexformer. For fair comparison, we plug the global MSA (G-MSA) used by previous CNN-Transformer hybrid methods into each basic unit of $\mathcal{R}$. The input feature maps of G-MSA are downscaled into $\frac{1}{4}$ size to avoid out of memory. We also compare our IG-MSA with local window-based MSA (W-MSA) proposed by Swin Transformer~\cite{liu2021swin}. As listed in Tab.~\ref{tab:msa}, our IG-MSA surpasses G-MSA and W-MSA by 1.41 and 1.34 dB while costing 2.08G and 0.86G FLOPS less. These results demonstrate the cost-effectiveness advantage of the proposed IG-MSA.

\vspace{-2mm}
\section{Conclusion}
\vspace{-1mm}
In this paper, we propose a novel Transformer-based method, Retinexformer, for low-light image enhancement. We start from the Retinex theory. By analyzing the corruptions hidden in the under-exposed scenes and caused by the light-up process, we introduce perturbation terms into the original Retinex model and formulate a new Retinex-based framework, ORF.  Then we design an IGT that utilizes the illumination information captured by ORF to direct the modeling of long-range dependences and interactions of regions with different lighting conditions. Finally, our Retinexformer is derived by plugging IGT into ORF. Extensive quantitative and qualitative experiments show that our Retinexformer dramatically outperforms SOTA methods on thirteen datasets. The results of user study and low-light detection also demonstrate the practical values of our method.

\vspace{1mm}

\noindent\textbf{Acknowledgements:} This research was funded through National Key Research and Development Program of China (Project No. 2022YFB36066), in part by the Shenzhen Science and Technology Project under Grant (CJGJZD2020 0617102601004, JCYJ20220818101001004) and Alexander von Humboldt Foundation.

{\small
\bibliographystyle{ieee_fullname}
\bibliography{reference}
}

\end{document}